\documentclass[sigconf]{acmart}
\AtBeginDocument{%
  }

\copyrightyear{2026}
\acmYear{2026}
\setcopyright{cc}
\setcctype{by}
\acmConference[CIKM '26]{Proceedings of the 35th ACM International Conference on Information and Knowledge Management}{November 07--11, 2026}{Rome, Italy}
\acmBooktitle{Proceedings of the 35th ACM International Conference on Information and Knowledge Management (CIKM '26), November 07--11, 2026, Rome, Italy}
\acmDOI{10.1145/3799682.3840912}
\acmISBN{979-8-4007-2539-5/2026/11}

\usepackage{algorithm}
\usepackage{algorithmic}
\usepackage{amsmath}
\usepackage{graphicx} 
\usepackage{booktabs}
\usepackage{multirow}
\begin{document}

\title{LENS: Reliability-Aware Stock Prediction via LLM-Extracted Priors and News-Driven Semantic Graphs}

\author{Hao Zhou}
\email{hao.zhou@ed.ac.uk}
\affiliation{%
  \department{School of Informatics}
  \institution{The University of Edinburgh}
  \city{Edinburgh}
  \country{United Kingdom}}

\author{Tiejun Ma}
\email{tiejun.ma@ed.ac.uk}
\affiliation{%
  \department{School of Informatics}
  \institution{The University of Edinburgh}
  \city{Edinburgh}
  \country{United Kingdom}}

\author{Luis Felipe Costa Sperb}
\email{f.costasperb@kingston.ac.uk}
\affiliation{%
  \department{Kingston Business School}
  \institution{Kingston University}
  \city{London}
  \country{United Kingdom}}

\renewcommand{\shortauthors}{Hao Zhou, Tiejun Ma, and Luis Felipe Costa Sperb}

\begin{abstract}
Predicting stock returns requires modeling persistent company relations and time-varying market signals. Graph Neural Network (GNN) approaches combine static priors with dynamic relational graphs, but treat every extracted edge as a trustworthy binary fact, ignoring heterogeneous reliability across relational sources. We propose \textbf{LENS} (\textbf{L}LM-\textbf{E}xtracted priors and \textbf{N}ews-driven \textbf{S}emantics), a static--dynamic graph framework for stock return prediction and ranking that models relational reliability as a first-class signal across the pipeline. On the static side, we extract a multi-relational company graph from a Large Language Model and aggregate it over multiple runs with confidence weighting to downweight low-consistency edges. On the dynamic side, we build a news-driven semantic graph with exponential-decay temporal aggregation to capture changing market context under uneven coverage. A context-dependent gated fusion module rebalances the two graphs at the per-stock, per-day level. On S\&P 500, LENS improves AR@5 by 16.4\% and SR@5 by 22.6\% over the strongest baseline, and on NASDAQ 100 it achieves comparable improvements (15.5\% and 20.9\% respectively). Ablations isolate each component's contribution, and a human-validation study of sampled relations supports that the aggregation is grounded, providing evidence that reliability-aware LLM relations and news-based semantic graphs carry complementary information for graph-based financial prediction.
\end{abstract}



\maketitle

\section{Introduction}
Stock return prediction is a fundamental problem in quantitative investment, and inter-company relationships are widely recognised as a valuable signal beyond a stock's own historical features.
These relations live on different timescales: persistent structural ties such as supply chains \cite{chen2022multivariate} and industry classifications \cite{feng2019rsr} change slowly, while event-driven interactions \cite{tian2023geru} can shift overnight.
Graph Neural Networks (GNNs) have become the standard tool for propagating information across such relations \citep{chen2018incorporating, li2021ijcai}, yet their effectiveness is fundamentally bottlenecked by the quality of the underlying graph: rigid static priors ignore real-time events, while noisy dynamic graphs amplify fake correlations during message passing.
We argue that the root cause is not a shortage of relational evidence but the way it is used: existing methods treat every extracted edge as an equally trustworthy binary fact, with no mechanism to model the heterogeneous reliability of different relational sources across stocks, time, and extraction methods.
Therefore, we propose to model the relational reliability as a first-class signal.

Stock prediction methods relying on static priors illustrate the issue clearly.
Traditional sources such as the Global Industry Classification Standard (GICS) \citep{sawhney2021hypergraph} or knowledge bases such as WikiData \cite{feng2019rsr} are stable and consistent but coarse and slow to reflect evolving business relationships.
Recent work \cite{zhou2025zero} replaces these with relations extracted directly from a Large Language Model (LLM) via zero-shot prompting, producing richer and more granular graphs than predefined taxonomies.
However, LLM extraction is inherently probabilistic: different runs of the same prompt can yield inconsistent or hallucinated triples, and prior work takes the LLM output as a deterministic, binary edge set with limited direct validation of the resulting graph.
Such uncertainty propagates through message passing and can reduce downstream prediction quality.

Dynamic graphs also face a similar reliability issue.
Many studies build dynamic relations from historical price correlations \cite{tian2023geru, cheng2021modeling, yin2022graph}, but such graphs are noisy and largely redundant with the price-derived node features that GNNs already encode.
Financial news offers a more semantically grounded alternative, yet news coverage is highly uneven across companies and time.
Aggregating headlines over a short window leaves many stocks isolated, while a long window can reduce the signal-to-noise ratio due to outdated information.
As a result, news-driven graphs alone are unreliable, particularly for stocks with lower news coverage.

To address these limitations, we propose \textbf{LENS} (\textbf{L}LM-\textbf{E}xtracted priors and \textbf{N}ews-driven \textbf{S}emantics), a static-dynamic graph framework for stock return prediction and ranking that models relational reliability as a first-class signal across the whole pipeline.
On the static side, LENS extracts a multi-relational company graph by prompting an LLM whose knowledge cutoff precedes the evaluation period (mitigating information leakage), treats the extraction as a stochastic process repeated over multiple runs, and aggregates the resulting triples into a reliability-weighted edge set whose weights combine cross-run frequency with model-reported confidence.
On the dynamic side, LENS constructs a news-driven semantic graph with exponential-decay temporal aggregation and mutual top-$k$ sparsification, retaining recency without ignoring sparsely covered companies.
A context-dependent gated fusion module then adaptively rebalances the static and dynamic representations at the per-stock, per-day level, leaning more on the stable LLM prior when news evidence is weak and on the news-derived signal when coverage is rich.

We evaluate LENS on the S\&P~500 and NASDAQ~100 with rolling quarterly out-of-sample folds, followed by a direct human-validation study of the LLM-extracted static graph and a coverage-stratified robustness analysis.
In sum, our main contributions are:

\begin{itemize}
    \item \textbf{Reliability-weighted LLM company priors, validated end-to-end.} We formulate LLM relation extraction as a stochastic graph-construction process and aggregate five extraction runs into reliability-weighted multi-relational edges by combining frequency with model-reported confidence. We then validate the resulting graph both downstream, where LLM priors outperform GICS, WikiData, and news-co-occurrence alternatives, and directly,
    via a 120-triple human-validation study under three independent annotators with web-search adjudication (Fleiss' $\kappa{=}0.58$, moderate), which yields 91.1\% relaxed precision for the default Highrecall graph and a monotonic association between reliability bins and strict correctness (36.7\% / 50.0\% / 76.7\% on low/mid/high terciles).
    \item \textbf{Noise-controlled dynamic semantic graphs under uneven news coverage.} We use exponential-decay temporal aggregation to construct daily news-driven semantic graphs to address the heterogeneous coverage and reliability of news evidence across stocks and time. This design outperforms price-correlation and news-co-occurrence dynamic-graph baselines.
    \item \textbf{Reliability-aware gated fusion across two markets.} We introduce a context-dependent gating module that balances the relative reliability of the static LLM prior and the dynamic news graph at the stock-day level: leaning more on the static prior when news evidence is weak and more on the dynamic graph when coverage is rich. Across the S\&P~500 and NASDAQ~100, LENS improves ranking and portfolio metrics over strong sequence- and graph-based baselines, and reduces the MAE gap between high- and low-news-coverage stocks from 5.39\% to 3.18\%.

\end{itemize}

\section{Related Work}
\paragraph{Graph-Based Stock Prediction.} Graph neural networks have been widely adopted for stock forecasting to model cross-company dependencies.
Early studies typically rely on predefined relationships, such as industry taxonomies \cite{sawhney2021hypergraph}, supply chains \cite{chen2022multivariate}, shareholding patterns \cite{wang2023hatri}, and knowledge graphs like  \cite{sawhney2020deepemnlp, feng2019rsr}.
While stable, these priors often suffer from rigidity and outdated information. Recently, \citet{zhou2025zero} demonstrated the potential of extracting more granular relational priors from LLMs. However, such probabilistic extractions introduce uncertainty that remains unaddressed. Parallel to static priors, dynamic graphs infer time-varying relations from price correlations or historical features \cite{tian2023geru, cheng2021modeling, yin2021ijcnn}, but they are often affected by low signal-to-noise ratios. Although some works attempt to jointly model static and dynamic links \cite{WANG2022MGconv, zhao2023TKDE}, they typically rely on fixed fusion rules (e.g., additive or static gating), which fail to account for the fluctuating reliability of different data sources under volatile market conditions.

\paragraph{Financial NLP and Semantic Graphs.} Financial NLP has evolved from basic sentiment analysis \cite{liu2021finbert} to leveraging LLMs for sophisticated information extraction, such as FinGPT \cite{zhang2023fingpt} and Bloomberg-GPT \cite{wu2023bloomberggpt}. Despite the rich semantics of news-driven signals, they face severe spatiotemporal fragmentation: uneven media coverage often fails to model companies with sparse coverage and leads to isolated parts in graph. Our work, \textsf{LENS}, bridges these gaps by introducing a reliability-aware framework that extracts reliable signals from stochastic LLM priors and employs temporal decay aggregation to mitigate uneven news coverage, followed by an adaptive fusion mechanism for robust forecasting.

\section{Problem Formulation}
Following \citep{feng2019rsr}, we formulate the stock movement prediction task as a learning to rank problem. Let $S = \{s_1, s_2, \dots, s_N\}$ denote the set of $N$ target stocks. For each stock $s_i \in S$ on a given trading day $t$, we define its future 1-day log return as $y_i^{t+1}=\log\!\left(\frac{p_i^{t+1}}{p_i^t}\right)$, where $p_i^t$ is the closing price. 
The ground-truth ranking is based on the future returns: $s_i$ is ranked higher than $s_j$ iff $y_i^{t+1} > y_j^{t+1}$.
Our objective is to learn a scoring function, $f$, that uses information available up to day $t$ and outputs a real-valued score $\hat{y}_i^{t+1}$ for each stock $s_i$. This score is implemented as the predicted 1-day log return. Higher $\hat{y}_i^{t+1}$ implies higher expected return and thus a higher rank.

To learn this scoring function, our model leverages three sources of information. 
First, to capture stable, long-term company relationships, we construct a \textbf{weighted multi-relational prior graph} 
$ G_S=\big(S,\{A_S^{(r)}\}_{r\in\mathcal{R}}\big) $, where $\mathcal{R}$ denotes the set of relation types and each $A_S^{(r)}\in\mathbb{R}^{N\times N}$ is a weighted adjacency matrix for relation $r$. 
Second, to model event-driven dependencies, we construct a \textbf{news-derived dynamic semantic graph} $G_D^{t}=(S,A_D^{t})$, where $A_D^{t}\in\mathbb{R}^{N\times N}$ is a weighted adjacency matrix in which the value quantifies \emph{semantic relatedness} between daily news representations of companies at time $t$. 
Finally, the intrinsic state of each stock is captured by daily feature matrices $\mathbf{X}^{t}\in\mathbb{R}^{N\times F_{\text{in}}}$ derived from historical market signals over a lookback window of length $T$. 

Formally, our task is to learn a scoring function $f$ that maps these heterogeneous and temporally distinct inputs to a vector of ranking scores:
\begin{equation}
\label{eq:problem_formulation}
\hat{\mathbf{y}}^{t+1}
= f\!\left(
G_S,\ G_D^{t},\ \{\mathbf{X}^{u}\}_{u=t-T+1}^{t}
\right) . 
\end{equation}

\section{Methodology}
\begin{figure*}[t]
  \centering
  \includegraphics[width=0.9\textwidth]{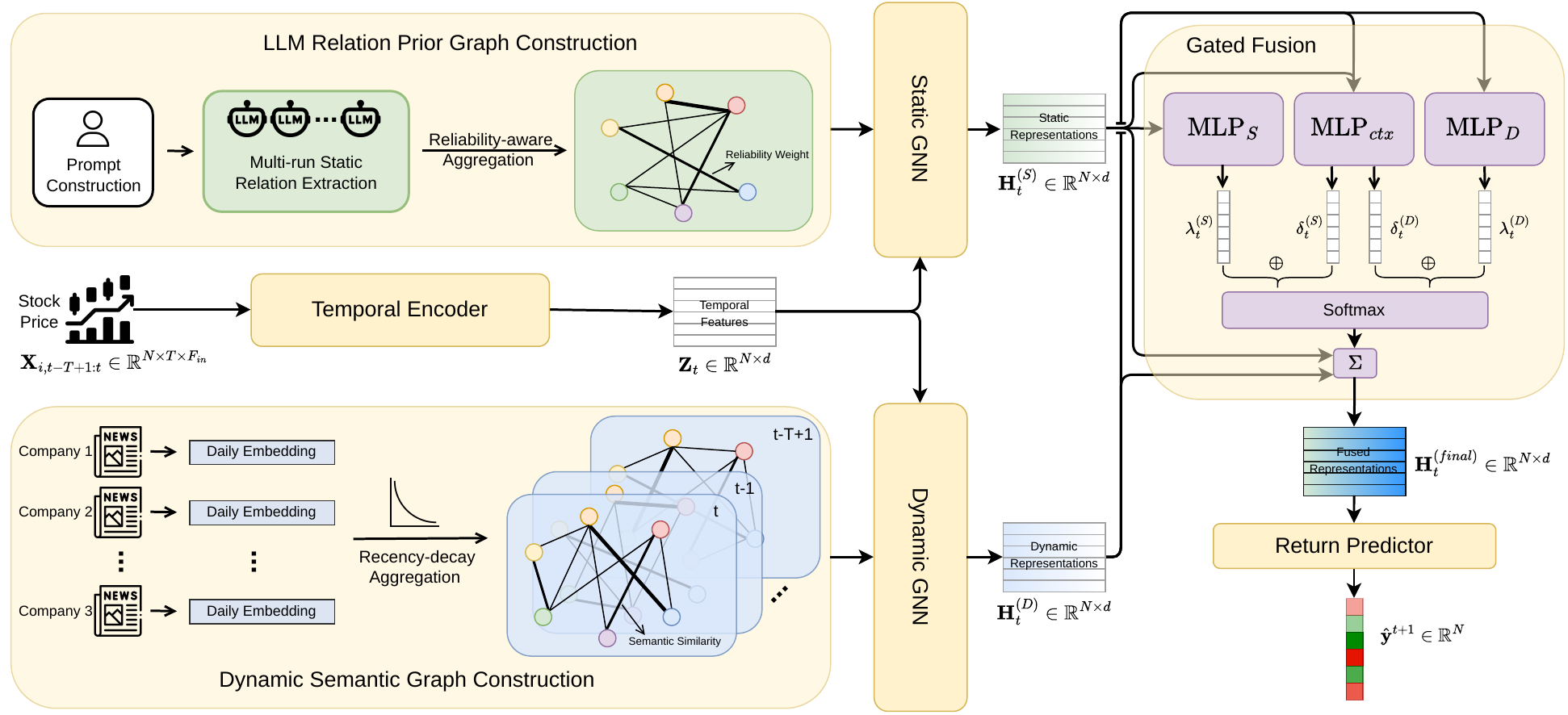}
  \caption{Overview of LENS. The framework consists of (1) static relation prior construction via multi-run LLM extraction and reliability-aware aggregation, (2) dynamic semantic graph construction from temporally aggregated news embeddings, (3) a dual-component GNN encoder over the static and dynamic graphs, and (4) a gated fusion module followed by a return predictor.}
  \Description{End-to-end pipeline of LENS: market features and news are encoded; a static multi-relational company graph is extracted by an LLM and aggregated across runs with reliability weights; a dynamic weighted semantic graph is built from time-windowed news embeddings; two graph encoders produce static and dynamic node representations; a gated fusion combines them to output next-day return/ranking scores.}
  \label{fig:model}
\end{figure*}
\subsection{Overview}
As illustrated in Figure~\ref{fig:model}, our architecture consists of the following main stages: (1) LLM-prompted static graph construction module, (2) news-driven dynamic semantic graph construction module, (3) a dual-component GNN encoder, (4) stock return \& ranking predictor.
This design allows the model to form a holistic and temporally-aware view of the market environment.

\subsection{Reliability-Weighted LLM Relation Priors}
\label{sec:reliLLM}
This module builds a static, typed relationship graph over constituent firms using a large language model (LLM). The static graph is designed as a slow-moving structural anchor: it is extracted once before the evaluation horizon and held fixed throughout testing, while fast-moving updates are absorbed by the dynamic graph (Section~\ref{sec:dynamic-graph}). To prevent leakage from LLM pretraining, the prompt restricts the evidence window to information available on or before 2023-09-30, matching the cutoff after which all test folds begin. Given a source company $i$, we prompt the LLM to output directed relation triplets $(i,r,j)$, where $j$ is also a constituent and $r$ belongs to a predefined taxonomy.

\paragraph{Relation taxonomy and output}
Following \citet{zhou2025zero}, we use 14 relationship types covering ownership (\textit{Subsidiary}, \textit{Parent}), investment (\textit{Investor}, \textit{InvestedBy}), acquisition (\textit{Acquirer}, \textit{Target}), partnership (\textit{JointVenture}, \textit{Strategic}, \textit{R\&D}, \textit{Collaborator}), commercial (\textit{Supplier}, \textit{Customer}, \textit{Distributor}), and competitive (\textit{Competitor}) ties.
We iterate over the S\&P 500 constituents and prompt the LLM to return up to 10 target tickers together with a scalar confidence score in [0,1] for each relation type.
We define the confidence score as the model's belief that the labeled relation is factually correct and publicly documented as of the cutoff date. The score is not intended to measure how close, important, or economically strong the connection is. This distinction is important because we later use confidence as a reliability signal when aggregating multiple extraction runs and when weighting edges in downstream learning.

\paragraph{Multi-run extraction and reliability-aware aggregation}
LLM relation extraction is stochastic, where repeated calls with the same prompt can yield different edge sets. We therefore treat $M$ independent runs as samples from an extraction process and aggregate them to improve stability.
For each triplet $e=(i,r,j)$ that appears, we count its frequency $f(e)$ and calculate its average confidence score $\bar{c}(e)$ over the runs where it appears. We then combine these two signals into an edge reliability weight:
\begin{equation}
    \rho(e) = f(e) \cdot \bar{c}(e) .
\end{equation}
In this way, we can mitigate the fluctuations in confidence scores that may be generated by different extraction runs and reduce the weight of those relations that appear by chance. Finally, in constructing a static relation graph, we propose three different aggregation strategies: (i) \textbf{Strict}, which retains an edge if its empirical frequency $f(e)\ge \theta_f$; (ii) \textbf{Highrecall}, which retains an edge if $f(e)\ge \theta_f$ \emph{or} its maximum confidence across runs $c_{\max}(e)>\theta_c$; (iii) \textbf{All}, which retains all extracted edges. The full procedure is summarised in Algorithm~\ref{alg:reliLLM}.

\begin{algorithm}[!t]
\caption{Reliability-Weighted LLM Relation Priors}
\label{alg:reliLLM}
\begin{algorithmic}[1]
\REQUIRE constituent firms $\mathcal{V}$; relation types $\mathcal{R}$; $M$ extraction runs; mode $\in\{\textsc{Strict},\textsc{Highrecall},\textsc{All}\}$
\ENSURE Static relation prior edges $\mathcal{E}_S$ with reliability weights $\rho(e)$
\STATE Set frequency threshold $\theta_f\leftarrow 0.6$ and confidence threshold $\theta_c\leftarrow 0.8$
\STATE Initialize counts $n(e)\leftarrow 0$, sums $s(e)\leftarrow 0$, max confidence $c_{\max}(e)\leftarrow 0$
\FOR{each source firm $i\in\mathcal{V}$ and run $m=1,\dots,M$}
  \STATE Query LLM; parse outputs into triplets $e=(i,r,j)$ with $c\in[0,1]$
  \FOR{each extracted $(e,c)$}
    \STATE $n(e)\mathrel{+}=1$;\ $s(e)\mathrel{+}=c$;\ $c_{\max}(e)\leftarrow\max(c_{\max}(e),c)$
  \ENDFOR
\ENDFOR
\STATE $\mathcal{E}_S\leftarrow\emptyset$
\FOR{each $e$ with $n(e)>0$}
  \STATE $f(e)\leftarrow n(e)/M$;\ $\bar{c}(e)\leftarrow s(e)/n(e)$;\ $\rho(e)\leftarrow f(e)\bar{c}(e)$
  \IF{mode $=$ \textsc{All}; \textbf{or} (\textsc{Strict} \AND $f(e)\!\ge\!\theta_f$); \textbf{or} (\textsc{Highrecall} \AND ($f(e)\!\ge\!\theta_f$ \OR $c_{\max}(e)\!>\!\theta_c$))}
    \STATE $\mathcal{E}_S\leftarrow\mathcal{E}_S\cup\{e\}$
  \ENDIF
\ENDFOR
\RETURN $\mathcal{E}_S$
\end{algorithmic}
\end{algorithm}

\subsection{Noise-Controlled Dynamic Semantic Graph from News}
\label{sec:dynamic-graph}

Whereas the static side addresses reliability at its source, the dynamic side faces a different problem: news coverage is uneven across stocks and across days, so any single-day similarity graph mixes informative signal with coverage noise. We address this with two complementary design choices: exponential-decay temporal aggregation, which balances recency against single-day fluctuations, and mutual top-$k$ sparsification with a targeted fallback for otherwise-isolated nodes, which suppresses pseudo-edges from accidental semantic overlap. The construction is described below.

\paragraph{Semantic encoding of news headlines} For each firm $i$ on day $t$, we collect its headline set $\mathcal{H}_{i}^{t}$ and concatenate them into a long sequence, then use a text embedding model to obtain a dense representation $\mathbf{e}_{i}^{t}$.

\paragraph{Temporal aggregation into window embeddings.}
We aggregate company-level daily news embedding into a window embedding of length $\tau$ over time mainly based on the following considerations: (i) The impact of major events to the market is often persistent. Using only the news of one day can easily ignore the lag effect of events and be more sensitive to single-day noise. Window aggregation can improve the stability and robustness of the representation while retaining key signals. (ii) News coverage is highly uneven across companies, with some companies having very little or no daily news. Window aggregation can significantly mitigate the ``isolated nodes'' problem caused by sparsity, thereby enabling information to flow fully on the graph. 
Specifically, we use an exponential decay weighting strategy:
\begin{equation}
\mathbf{e}_{i, t}^{(\mathrm{win})}
=\sum_{k=0}^{\tau-1}
\frac{\exp\!\left(-\alpha k\right)}
{\sum_{l=0}^{\tau-1}\exp\!\left(-\alpha l\right)}
\,\mathbf{e}_{i}^{t-k} , 
\end{equation}
where a higher $\alpha$ yields a stronger decay of the historical information. We treat $\tau$ and $\alpha$ as key design factors and analyze their impact in experiments.

\paragraph{Similarity-based edge construction.}
Given window embeddings $\{\mathbf{e}_{i, t}^{(\mathrm{win})}\}$, we compute semantic relatedness using cosine similarity:
\begin{equation}
\label{eq:dyn_sim}
s_{ij,t}=\cos(\mathbf{e}_{i, t}^{(\mathrm{win})},\mathbf{e}_{j, t}^{(\mathrm{win})}).
\end{equation}
Equation~(\ref{eq:dyn_sim}) defines a dense pairwise similarity matrix as an intermediate object; the dynamic GNN graph itself is \emph{not fully connected}. It is constructed by retaining only mutual top-$k_1$ neighbors: for each node $i$, we form its top-$k_1$ nearest-neighbor set $\mathcal{N}_{k_1}(i)$, and edge $(i,j)$ is included iff $j \in \mathcal{N}_{k_1}(i)$ and $i \in \mathcal{N}_{k_1}(j)$.
This mutual constraint suppresses unidirectional pseudo-connections caused by noisy news, short headlines, or accidental semantic overlap, and avoids the over-smoothing and noise propagation associated with overly dense graphs.
News coverage is long-tailed, however, so mutual-$k_1$ sparsification can leave some nodes isolated; we therefore add top-$k_2$ neighbors \emph{only} for such isolated nodes as a minimum-connectivity fix, preserving message-passing usability without degrading edge quality. The full construction is summarised in Algorithm~\ref{alg:dynamic_graph}.

\begin{algorithm}[!t]
\caption{News-Derived Dynamic Semantic Graph}
\label{alg:dynamic_graph}
\begin{algorithmic}[1]
\REQUIRE Daily news $\{\mathcal{H}_i^t\}$ for $i\in\mathcal{V}$; window $\tau$; decay $\alpha$; mutual top-$k_1$; fallback $k_2$
\ENSURE Daily dynamic graph $G_D^t=(\mathcal{V},\mathcal{E}_D^t,w_D^t)$
\STATE Embed $\mathbf{e}_i^t\leftarrow\mathrm{Voyage}(\mathcal{H}_i^t)$ for each firm $i\in\mathcal{V}$
\STATE $\mathbf{e}_{i,t}^{(\mathrm{win})}\!\leftarrow\!\sum_{k=0}^{\tau-1}\pi_k\,\mathbf{e}_i^{t-k}$,\ $\pi_k\!\propto\!\exp(-\alpha k)$
\STATE $s_{ij,t}\leftarrow\cos(\mathbf{e}_{i,t}^{(\mathrm{win})},\mathbf{e}_{j,t}^{(\mathrm{win})})$ for all $(i,j)$
\STATE $\mathcal{N}_{k_1}(i)\leftarrow$ top-$k_1$ neighbours of $i$ by $s_{ij,t}$; $\mathcal{E}_D^t\leftarrow\emptyset$
\FOR{each pair $(i,j)$ with $j\!\in\!\mathcal{N}_{k_1}(i)$ \AND $i\!\in\!\mathcal{N}_{k_1}(j)$}
  \STATE Add $(i,j)$ to $\mathcal{E}_D^t$ with $w_{ij,t}^D\leftarrow\max(s_{ij,t},0)$
\ENDFOR
\STATE \textbf{Fallback:} for each isolated $i$, attach its top-$k_2$ neighbours by $s_{ij,t}$ with weights $\max(s_{ij,t},0)$
\RETURN $G_D^t$
\end{algorithmic}
\end{algorithm}

We note that headline-level semantic embeddings are coarser than explicit event or entity-relation extraction and can admit topical co-mentions. The framework is designed to be robust to such noise: the reliability-weighted static prior (Section~\ref{sec:reliLLM}) provides a stable structural anchor, and the reliability-aware gated fusion (Section~\ref{sec:fusion}) adaptively downweights the dynamic graph when news signals are weak or sparse. Finer-grained event extraction is a natural extension that we leave to future work.

\subsection{Temporal Encoder}
\label{sec:temporal-encoder}

For each company $i$ and trading day $t$, we firstly form a length-$T$ sequence of per-day inputs (e.g., price and derived technical features) $\mathbf{X}_{i,t-T+1:t}\in\mathbb{R}^{T\times F_{in}}$. We feed this sequence into a standard Transformer with $H$ attention heads to capture temporal dependencies for each company individually. By stacking all companies we obtain $\mathbf{Z}_t\in\mathbb{R}^{N\times d}$, which is used as input to both graph encoders (Section~\ref{sec:dual-gnn}).
\begin{equation}
\mathbf{Z}_t = \mathrm{Transformer}\left(\{\mathbf{X}_{i,t-T+1:t}\}_{i=1}^{N}\right) . 
\end{equation}

\subsection{Dual Component GNN Encoder}
\label{sec:dual-gnn}
We then encode relational information with two parallel graph encoders: a \emph{static} encoder on the LLM-derived relationship prior $G_{S}$ and a \emph{dynamic} encoder on the news-driven semantic graph $G_D^{t}$. We keep the two encoders separate and only fuse their outputs at the end of the GNN block. This design reflects the different roles of the graphs: $G_{S}$ provides stable structural priors, while $G_D^{t}$ is time-varying and can be noisy under sparse news coverage.

\paragraph{Static encoder: Union Graph Transformer.}
The static graph $\mathcal{G}_S$ contains relation types $r_{ij}$ and reliability weights $\rho_{ij}$. We follow the Graph Transformer paradigm \cite{dwivedi2021graph} and inject edge features into attention to distinguish relation semantics and suppress low-reliability edges.
This design avoids the cost of maintaining a separate subgraph for each relation.
For each head $h$, we first compute
\begin{equation}
\mathbf{q}_{i,t}=\mathbf{W}_Q \mathbf{z}_{i,t}, \ \mathbf{k}_{j,t}=\mathbf{W}_K \mathbf{z}_{j,t},\ \mathbf{v}_{j,t}=\mathbf{W}_V \mathbf{z}_{j, t} . 
\end{equation}

Then, edge features are constructed using relation embedding and reliability scaling.
\begin{equation}
\mathbf{e}_{ij}=\rho_{ij}\cdot \mathbf{W}_{\text{rel}}[r_{ij}],
\end{equation}
Here $\rho_{ij}\in[0,1]$ scales the learnable relation embedding by the edge's reliability. $\mathbf{e}_{ij}$ is injected into both the key and value vector:
\begin{equation}
\label{eq:beta}
\beta_{ij,t} = \frac{\langle \mathbf{q}_{i,t},\; \mathbf{k}_{j,t} + \mathbf{e}_{ij}\rangle}{\sqrt{d_h}},
\end{equation}
\begin{equation}
\alpha_{ij,t} = \mathrm{Softmax}_{j\in\mathcal{N}_S(i)}(\beta_{ij,t}),
\end{equation}
\begin{equation}
\label{eq:value}
\mathbf{h}^{(S)}_{i,t}=\sum_{j\in\mathcal{N}_S(i)} \alpha_{ij,t}\,\big(\mathbf{v}_{j,t} + \mathbf{e}_{ij}\big).
\end{equation}
where $\mathbf{W}_Q, \mathbf{W}_K, \mathbf{W}_V, \mathbf{W}_{rel}$ are all independently learnable parameters for the head.
Finally, we concatenate the output vectors of each attention head along the feature dimension to obtain the final representation of the node.
We inject $\mathbf{e}_{ij}$ into both the key (Eq.~\ref{eq:beta}) and the value (Eq.~\ref{eq:value}) simultaneously: key-side injection modulates \emph{which} neighbors receive attention, while value-side injection modulates \emph{what} information is aggregated. Our ablation in Section~\ref{sec:rq4} confirms that dual injection outperforms either single-injection variant.

\paragraph{Dynamic encoder: Dynamic Graph Transformer.}
Unlike static relationship graphs, dynamic graphs involve single relationship weighted by semantic similarity (Section~\ref{sec:dynamic-graph}). 
Following Graphormer \cite{ying2021graphomer}, we employ a structural bias injection approach and incorporate the edge weight as a scalar bias term in the attention logits.
For each edge $(i,j)$ with weight $w^{D}_{ij,t}$,
\begin{equation}
\gamma_{ij,t}=\frac{\langle \mathbf{q}_{i,t},\; \mathbf{k}_{j,t}\rangle}{\sqrt{d_h}} + \log(1+w^{D}_{ij,t}),
\end{equation}
\begin{equation}
\alpha_{ij,t}=\mathrm{Softmax}_{j\in\mathcal{N}_D^t(i)}(\gamma_{ij,t}),
\end{equation}
\begin{equation}
\mathbf{h}^{(D)}_{i,t}=\sum_{j\in\mathcal{N}_D^t(i)} \alpha_{ij,t}\, \mathbf{v}_{j,t}.
\end{equation}

\paragraph{Stacking, residuals, and normalization.}
Both encoders can stack for $l_S$ and $l_D$ layers. After each layer, we apply a residual connection with a projection when dimensions change, followed by layer normalization; ReLU and dropout are applied between layers except the last. 

\subsection{Reliability-Aware Gated Fusion of Static and Dynamic Representations}
\label{sec:fusion}
To synthesize $\mathbf{H}_S$ from the static encoder and $\mathbf{H}_D$ from the dynamic encoder, we propose an interpretable residual gating mechanism inspired by Mixture of Experts (MoE) systems \cite{shazeer2017moe} and Residual Learning \cite{he2016deep}. The gate is the final stage at which relational reliability is rebalanced: when news evidence is weak (e.g., low-coverage stocks) the gate leans on the static prior, whereas under rich news coverage it relies more on the news-driven dynamic graph.
Instead of naive concatenation and projection, we decompose the gating decision into intrinsic assessment and contextual correction to enhance optimization stability. 

First, we use two independent MLPs to produce an intrinsic gate logit for each branch, capturing how informative the static and dynamic representations are on their own (before considering their interaction):
This decoupled design ensures a direct gradient path for each encoder, facilitating efficient optimization:
\begin{equation}
\lambda_{i,t}^{(S)} = \text{MLP}_S(\mathbf{h}_{i,t}^{(S)}), \quad \lambda_{i,t}^{(D)} = \text{MLP}_D(\mathbf{h}_{i,t}^{(D)}) . 
\end{equation}

To capture cross-graph dependencies, we concatenate the representations to form a global context, then employ a context-aware network to learn a correction term rather than predicting the gate weights from scratch:
\begin{equation}
[\delta_{i,t}^{(S)}, \delta_{i,t}^{(D)}] = \text{MLP}_{ctx}(\mathbf{h}_{i,t}^{(S)} \mathbin\Vert \mathbf{h}_{i,t}^{(D)} ) , 
\end{equation}
where $\mathbin\Vert$ denotes concatenation. The term $\delta$ adjusts the intrinsic gate logits based on the interaction between static and dynamic features.

The final weights $\alpha$ are obtained via softmax over the corrected scores ($\lambda + \delta$), producing the fused representation $\mathbf{H}^{(final)}$:
\begin{equation}
\mathbf{h}_{i,t}^{(final)} = \text{LayerNorm}\left( \sum_{k \in {S, D}} \text{Softmax}(\lambda_{i,t}^{(k)} + \delta_{i,t}^{(k)}) \cdot \mathbf{h}_{i,t}^{(k)} \right) . 
\end{equation}

\subsection{Stock Return \& Ranking Predictor}

For each stock $s_i$ on day $t$, the fused representation $\mathbf{h}_{i,t}^{\text{final}}$ is fed into a two-layer MLP predictor to produce a real-valued score $\hat{y}_{i}^{t+1}$: 
\begin{equation}
\hat{y}_{i}^{t+1} = \mathbf{W}_2\mathrm{ReLU}(\mathbf{W}_1 \mathbf{h}_{i,t}^{\text{final}} + b_1)  + b_2 , 
\end{equation} 
where $\mathbf{W}_1\in\mathbb{R}^{\frac{d}{2}\times d}$, $\mathbf{W}_2\in\mathbb{R}^{1\times \frac{d}{2}}$, and $b_1,b_2$ are learnable parameters.

\subsection{Training Objective}
To simultaneously optimize for return prediction and ranking, we employ a hybrid objective function that linearly combines Mean Squared Error (MSE) and pairwise RankNet loss:
\begin{equation}
\mathcal{L}=\mathcal{L}_{\text{mse}}+\phi\,\mathcal{L}_{\text{rank}}.
\end{equation}
\paragraph{Regression loss} We use the standard MSE loss to minimize the deviation between predicted and ground-truth returns:
\begin{equation}
\mathcal{L}_{\text{mse}}=\frac{1}{N}\sum_{i=1}^{N}\left(\hat{y}^{t+1}_i-y^{t+1}_i\right)^2.
\end{equation}
\paragraph{Ranking loss} To capture ordinal relationships, we utilize RankNet \cite{burges2005ranknet}. The probability that company $i$ outperforms company $j$ is modeled via a logistic function of their predicted score difference:
\begin{equation}
    P_{i,j} = \frac{1}{1 + \exp\left(-\tau (\hat{y}^{t+1}_i - \hat{y}^{t+1}_j)\right)} , 
\end{equation} 
where $\tau$ is a scaling hyperparameter controlling the steepness of the sigmoid curve. 

The ground truth probability $\bar{P}_{i,j}$ is defined as: $\bar{P}_{i,j} =\mathbb{I}(y^{t+1}_i>y^{t+1}_j)$.
To enhance training stability and mitigate the impact of negligible return differences (noise), we introduce a threshold $\epsilon$. We only consider valid pairs $\Omega = \{(i, j) \mid |y^{t+1}_i - y^{t+1}_j| > \epsilon\}$. The ranking loss is computed as the binary cross-entropy over all valid pairs:
\begin{equation}
    \mathcal{L}_{\text{rank}} = - \frac{1}{|\Omega|} \sum_{(i,j) \in \Omega} \left[ \bar{P}_{i,j} \log P_{i,j} + (1 - \bar{P}_{i,j}) \log (1 - P_{i,j}) \right] . 
\end{equation}

\section{Experiments}
Our experimental study is organised around the reliability-as-first-class-signal claim made in Sections~\ref{sec:reliLLM}--\ref{sec:fusion}. We probe this claim from four angles: (i) does it produce competitive ranking and portfolio performance across markets; (ii) is the reliability-weighted LLM-extracted graph itself empirically high-quality; (iii) does the news-driven dynamic graph contribute complementary signal; and (iv) does the reliability-aware gated fusion deliver the predicted robustness for stocks with weak news coverage.
\subsection{Key Research Questions}

\textbf{RQ1:} How does LENS perform compared to existing sequential and graph-based methods on cross-sectional ranking and portfolio metrics, and do the gains transfer across market universes?

\noindent\textbf{RQ2:} Is the reliability-weighted LLM-extracted static graph empirically high-quality, and do LLM-extracted priors with reliability-aware aggregation outperform alternative static graph sources?

\noindent\textbf{RQ3:} Does the news-driven dynamic semantic graph add complementary signal beyond the static prior, and how do its temporal aggregation parameters affect graph quality?

\noindent\textbf{RQ4:} Does the reliability-aware gated fusion improve robustness, particularly for stocks with sparse news coverage?

\subsection{Datasets}
\paragraph{Markets and features.}
We evaluate LENS on two US large-cap equity universes: \textbf{S\&P 500} and \textbf{NASDAQ 100}, both using constituent lists fixed on September 30, 2023 and filtered to require complete daily trading data over the full horizon (450 and 87 stocks respectively). For each stock we collect daily OHLCV and standard technical indicators (MA, RSI, MACD)~\cite{kumbure2022machine}, normalised into the per-stock daily feature vector $\mathbf{x}_i^t$. The two universes share evaluation windows, baselines, feature pipeline, and news source; only the constituent list and corresponding news subset differ.

\paragraph{News corpus.}
For dynamic graph construction, we use the Thomson Reuters News Archive restricted to each universe's constituents. TRNA's \texttt{assetId} metadata maps each item to one or more constituents, so multi-firm items are assigned to every tagged firm without matching company names in the headline. We focus on headlines, which carry informative event signals for short-horizon prediction~\cite{ding2014using, vargas2017deep}, and embed daily headlines with Voyage-3-large, a text embedding model that performs strongly on the FinMTEB financial benchmark~\cite{tang2025finmteb}.

\paragraph{Splits and leakage mitigation.}
\label{para:leakage}
We employ rolling-window cross-validation with five folds: each fold uses approximately 7 years of training data, a 3-month validation period, and a 3-month test period, with test windows rolling forward quarterly (Oct--Dec 2023 through Oct--Dec 2024)~\cite{gu2020empirical}. LLM-extracted relational graphs introduce a non-trivial leakage risk, which we mitigate in three places: (i) the extractor's knowledge cutoff (o3-mini, 2023-10-01)\footnote{\url{https://platform.openai.com/docs/models/o3-mini}. 2023-10-01 denotes the model's knowledge cutoff rather than its release date} bounds the static prior; (ii) the extraction prompt restricts the admissible evidence window to information available on or before 2023-09-30; (iii) all five test folds begin on or after October 2023, so any post-cutoff drift between the static graph and the market is itself part of the test signal. The dynamic graph uses only headlines published on or before day $t$ when predicting day $t+1$.
\begin{table*}[t!]
    \centering
    \caption{Main performance comparison on S\&P 500 and NASDAQ 100. We report the mean over 10 seeds. For each metric within each market, the best result is in \textbf{bold} and the second-best is \underline{underlined}. $\uparrow$ indicates higher is better. AR and SR are reported with a 5\,bps transaction cost. ** denotes $p<0.01$ and * denotes $p<0.05$ via paired t-test over 10 seeds.}
    \label{tab:main_results}
    \setlength{\tabcolsep}{2.5pt}
    \scalebox{0.85}{
    \begin{tabular}{l ccccccc @{\hspace{1.2em}} ccccccc}
        \toprule
        & \multicolumn{7}{c}{\textbf{S\&P 500}} & \multicolumn{7}{c}{\textbf{NASDAQ 100}} \\
        \cmidrule(lr){2-8} \cmidrule(lr){9-15}
        & \textbf{MRR}$\uparrow$ & \textbf{AR@5}$\uparrow$ & \textbf{SR@5}$\uparrow$ & \textbf{AR@10}$\uparrow$ & \textbf{SR@10}$\uparrow$ & \textbf{AR@20}$\uparrow$ & \textbf{SR@20}$\uparrow$
        & \textbf{MRR}$\uparrow$ & \textbf{AR@5}$\uparrow$ & \textbf{SR@5}$\uparrow$ & \textbf{AR@10}$\uparrow$ & \textbf{SR@10}$\uparrow$ & \textbf{AR@20}$\uparrow$ & \textbf{SR@20}$\uparrow$ \\
        \midrule
        o3-mini      & 0.0232 & 0.2501 & 1.0407 & 0.1534 & 0.8023 & 0.1617 & 0.9481 & 0.0638 & 0.1872 & 0.7890 & 0.1549 & 0.7436 & 0.1206 & 0.7351 \\
        FinGPT       & 0.0362 & 0.3997 & 1.4424 & 0.3172 & 1.3423 & 0.2772 & 1.3100  & 0.0852 & 0.3127 & 1.0853 & 0.2104 & 0.9010 & 0.2098 & 1.0394 \\
        \midrule
        MLP          & 0.0359 & 0.3564 & 1.1679 & 0.3128 & 1.2111 & 0.2585 & 1.1595 & 0.0859 & 0.2734 & 0.8419 & 0.1953 & 0.7826 & 0.2167 & 0.9128 \\
        ALSTM        & 0.0275 & 0.2401 & 1.0153 & 0.2619 & 1.2322 & 0.2663 & 1.3107 & 0.0712 & 0.1684 & 0.7128 & 0.1745 & 0.8273 & 0.2089 & \underline{1.0431} \\
        Transformer  & 0.0364 & 0.3287 & 1.2402 & 0.3235 & 1.4030 & 0.2628 & 1.2959 & \underline{0.0915} & 0.2543 & 0.8957 & 0.2127 & 0.9381 & \underline{0.2178} & 1.0250 \\
        StockMixer   & \underline{0.0365} & 0.2855 & 1.1607 & 0.2617 & 1.3026 & 0.2265 & 1.2890 & 0.0900 & 0.2218 & 0.8347 & 0.1742 & 0.8492 & 0.1856 & 1.0387 \\
        \midrule
        RSR-I        & 0.0286 & 0.1901 & 0.8715 & 0.2158 & 1.1280 & 0.2007 & 1.1959 & 0.0738 & 0.1376 & 0.6231 & 0.1428 & 0.7593 & 0.1672 & 0.9531 \\
        RTGCN        & 0.0269 & 0.1695 & 0.7757 & 0.1489 & 0.7695 & 0.1454 & 0.8432 & 0.0750 & 0.1324 & 0.5847 & 0.0935 & 0.5184 & 0.1102 & 0.6735 \\
        GERU         & 0.0290 & 0.1424 & 0.5765 & 0.1516 & 0.7189 & 0.1450 & 0.7797 & 0.0697 & 0.1083 & 0.4391 & 0.1215 & 0.5232 & 0.1342 & 0.7211 \\
        MASTER       & 0.0281 & 0.2888 & 1.2085 & 0.2317 & 1.1769 & 0.2196 & 1.2172 & 0.0726 & 0.2231 & 0.9023 & 0.1556 & 0.7912 & 0.1731 & 0.9712 \\
        COGRASP      & 0.0267 & \underline{0.4493} & \underline{1.5249} & \underline{0.3404} & \underline{1.4969} & \underline{0.2927} & \underline{1.4088} & 0.0742 & \underline{0.3411} & \underline{1.1357} & \underline{0.2218} & \underline{0.9637} & 0.2089 & 1.0276 \\
        \midrule
        \textbf{LENS} & \textbf{0.0371} & \textbf{0.5232**} & \textbf{1.8694**} & \textbf{0.3826*} & \textbf{1.6238*} & \textbf{0.3018} & \textbf{1.4575*} & \textbf{0.0937} & \textbf{0.3939**} & \textbf{1.3725**}  & \textbf{0.2460*}  & \textbf{1.0710*} & \textbf{0.2427*} & \textbf{1.1818*} \\
        \bottomrule
    \end{tabular}}
\end{table*}

\subsection{Baselines}
We compare LENS against representative methods including sequential models and graph-based approaches. 
(1) \textbf{o3-mini}: the same LLM as used for extracting static relation priors in LENS, but we directly use it for return prediction.
(2) \textbf{FinGPT} \cite{zhang2023fingpt}: an LLM for financial sentiment analysis. We use it to predict the sentiment polarity of the news, which is then concatenated with price features for return prediction.
(3) \textbf{MLP} \cite{rumelhart1986learning}: a simple feedforward network that processes each stock's features independently, serving as a non-temporal, non-relational baseline. 
(4) \textbf{ALSTM} \cite{qin2017alstm}: an LSTM variant with an attention mechanism to capture salient temporal patterns. 
(5) \textbf{Transformer} \cite{vaswani2017attention} employs multi-head self-attention to model complex temporal dependencies. 
(6) \textbf{StockMixer} \cite{fan2024stockmixer}: an MLP-Mixer architecture that alternates mixing across time and stock dimensions, representing recent advances in non-graph stock prediction. 
(7) \textbf{RSR-I} \cite{feng2019rsr} constructs a static graph based on industry taxonomies and applies relational modeling via temporal graph convolution. 
(8) \textbf{RTGCN} \cite{zheng2023rtgcn} builds static graphs from industry taxonomies and Wikipedia and uses recurrent graph convolution to capture evolving features. 
(9) \textbf{GERU} \cite{tian2023geru} uses a dynamic graph neural network to automatically learn the evolving dependencies from historical stock features. 
(10) \textbf{MASTER} \cite{li2024master}: a market-guided stock transformer, which models the stock correlation implicitly and uses market information for automatic feature selection.
(11) \textbf{COGRASP} \cite{li2025cograsp} constructs graphs based on news co-occurrence patterns, capturing media-driven relationships between companies.

All methods share the same stock universes, temporal splits, targets, and evaluation protocol, but retain their original information sources. Table~\ref{tab:main_results} therefore compares end-to-end systems; the matched ablations in Sections~\ref{sec:rq2}--\ref{sec:rq4} isolate graph source, reliability weighting, and fusion.

\subsection{Evaluation Metrics}

We evaluate LENS from two complementary perspectives that align with our task objective: ranking quality and portfolio profitability. We adopt \textbf{Mean Reciprocal Rank (MRR)} to measure how well the model ranks the best-performing stock: for each trading day, we compute the reciprocal rank of the stock with the highest actual return within the predicted ranking and average over all days. 
To assess portfolio performance, we construct equal-weighted, daily-rebalanced long-only portfolios from the top-$K$ stocks, with $K\in\{5,10,20\}$, and deduct 5\,bps times one-way turnover (half the $L_1$ weight change). \textbf{Annualised Return (AR)} compounds net daily returns to a 252-day annual rate; \textbf{Sharpe Ratio (SR)} annualises daily excess-return mean and volatility using a 2\% annual risk-free rate.
MRR, AR, and SR are aggregate metrics and cannot characterise how accuracy varies across stock subsets. We therefore report \textbf{Mean Absolute Error (MAE)} in Table~\ref{tab:mae_tier}, where stocks are stratified by news-coverage tier, to assess robustness under sparse coverage.

\subsection{Implementation Details}
LENS and all baselines are implemented in PyTorch. All models are trained on a single NVIDIA L40 GPU. We repeat each experiment with 10 different random seeds and report the averaged results to ensure statistical robustness.\footnote{Seeds are derived deterministically from a global seed (default 42); cuDNN benchmark is disabled, deterministic algorithms and seeded RNGs are enforced for \texttt{random}, \texttt{numpy}, and \texttt{torch} (CPU+CUDA), and TF32 is disabled on Ampere/Hopper GPUs.} All baselines are tuned following their original papers. We use Adam optimizer with learning rate $1 \times 10^{-3}$, weight decay $1 \times 10^{-5}$. Dropout is set to 0.1, and batch size is 32. We train all the models for 100 epochs and apply early stopping with a patience of 20 epochs.

For the static graph, we perform $M=5$ independent extraction runs using o3-mini, which does not expose a user-controlled temperature; the prompt and all supported API settings are held fixed across runs. We adopt the Highrecall aggregation strategy.
For the dynamic graph, we set the temporal window $\tau=10$ days with decay rate $\alpha=0.5$. We use mutual top-$k_1=20$ for edge construction and top-$k_2=5$ for isolated node connectivity. The temporal encoder uses a lookback window of $T=20$ days. The hidden dimension is $d=128$, and the number of attention heads is 4 across all components. We use a single GNN layer for both static and dynamic encoders, and a 2-layer temporal encoder.

\subsection{Overall Performance Comparison (RQ1)}
\label{sec:rq1}

Table~\ref{tab:main_results} shows that LENS achieves the best reported ranking and portfolio metrics on the S\&P 500.
In particular, LENS achieves the highest MRR (0.0371) and the strongest trading performance, with annualised returns of 52.32\%, 38.26\%, and 30.18\% for Top-5, Top-10, and Top-20 portfolios, and corresponding Sharpe ratios of 1.87, 1.62, and 1.46.
Moreover, LENS shows statistically significant improvements over all baselines on most AR/SR metrics (marked by $*$ or $**$ in Table~\ref{tab:main_results}).
These backtest results indicate potential portfolio value under the stated assumptions.

The two LLM-based baselines show different ways of using news. FinGPT performs better than o3-mini, and both are inferior to LENS. The comparison between o3-mini and LENS suggests that treating LLMs as standalone predictors is less effective. Compared with the price-only MLP baseline, FinGPT suggests that incorporating an additional sentiment signal from news can provide modest benefits. However, the improvements are not significant in most metrics.

Among sequential baselines, Transformer and StockMixer show competitive MRR performance, but their portfolio performance is consistently weaker than LENS. Under their prescribed relation sources, static graph methods (RSR-I and RTGCN) do not consistently outperform sequence models, motivating the controlled analysis of graph construction and reliability in Section~\ref{sec:rq2}.
GERU, which constructs dynamic graphs based on historical price characteristics, performs poorly on most metrics, suggesting that dynamic relationships derived solely from historical prices are insufficient to capture profitable signals.
COGRASP, which constructs relation priors from news co-occurrence, is the strongest baseline on portfolio metrics. LENS improves over COGRASP by 16.4\% in AR@5 and 22.6\% in SR@5. The matched ablations suggest two contributors to this gap: replacing semantic edges with news co-occurrence reduces performance (Section~\ref{sec:rq3}), while removing either the static or dynamic graph also produces clear declines (Sections~\ref{sec:rq2}--\ref{sec:rq3}), supporting the complementarity of the two sources.
We also observe that all graph-based methods achieve low MRR, likely due to over-smoothing. They tend to mix neighbour information and weaken extreme signals in message passing.

To test whether the reliability-aware framework generalises beyond a single market universe, we replicate the protocol on NASDAQ 100 with identical baselines, features, and rolling-window schedule (right panel of Table~\ref{tab:main_results}). The relative ordering of methods is almost consistent with the S\&P 500 setting: LENS attains the best results on ranking and portfolio metrics, with the largest gains concentrated on small-$K$ portfolios. This suggests that the dominant performance is driven by the reliability-aware design rather than by S\&P 500-specific artefacts. 

Beyond mean comparison, we examine the consistency of LENS's gains on S\&P 500. Against the strongest baseline (COGRASP), LENS wins on AR@5 and SR@5 in \emph{all five} quarterly out-of-sample folds, including the most challenging Fold 4 (Jul-Sep 2024), which is a period spanning a sector rotation, a flash crash, and a Federal Reserve rate cut. Across 10 random seeds, LENS attains the highest AR@5 in 9 of 10 seeds and the highest SR@5 in 10 of 10. The win-rate framing complements the paired t-test markers in Table~\ref{tab:main_results} and confirms that the observed gains are not concentrated in a few favourable folds or seeds. 

\subsection{Direct Validation of LLM Relations (RQ2)}
\label{sec:human-validation}

Because LLM-extracted relational graphs are rarely evaluated directly and downstream prediction performance is the main proxy, we conduct a human validation of the static graph.

Specifically, we sample 120 candidate triples (90 from the LENS Highrecall graph, 30 from a single-run baseline). 
Three annotators independently web-search public evidence available on or before the cutoff and assign each triple one of three verdicts: \emph{supported} (real connection, exact label and direction), \emph{partial} (real connection, imperfect label or direction), or \emph{unsupported}; a final adjudicator resolves disagreements (80 unanimous, 30 majority, 10 three-way splits). We report \textbf{Strict} precision $= \#\text{supported}/n$ and \textbf{Relaxed} precision $= (\#\text{supported}+\#\text{partial})/n$. Pre-adjudication Fleiss' $\kappa = 0.58$ on the three-way labels.

Table~\ref{tab:human_val} shows the validation precision. 
Highrecall aggregation reaches \textbf{91.1\% relaxed precision} versus 83.3\% for the single-run baseline (Table~\ref{tab:human_val}), nearly halving the unsupported rate from 16.7\% to 8.9\%. Strict precision is 54.4\%, within sampling noise of single-run's 56.7\% on $n{=}30$ (two-proportion z-test $p{=}0.83$). The small gap is the expected cost of Highrecall's broader coverage, and Sec. \ref{sec:rq2} indicates that Highrecall is more advantageous in downstream metrics. The 37\% relaxed--strict gap therefore reflects label imprecision rather than fabrication: most triples in the gap have a real business connection but an approximate LLM-assigned relation type (e.g., `strategic partner' vs `supplier'). Restricting to $f(e)\geq 0.6$ raises relaxed precision to 92.9\% (strict 78.6\%) and non-competitor relations reach 96.2\% (strict 63.5\%). Note that LLM-reported confidence is monotonically associated with strict label correctness across terciles (36.7\% / 50.0\% / 76.7\%), so $\rho(e)$ carries per-edge reliability information the downstream model uses directly as an edge weight.

Two observations are important for the following ablations. First, although strict per-edge precision is moderate,  the monotonic association means that $\rho$-weighting can systematically suppress less-supported edges in message passing without requiring a hard binary threshold.
Secondly, there is a trade-off between recall and precision: aggressive frequency thresholding improves precision but discards low-frequency edges that, while individually uncertain, collectively encode useful cross-company structures.
This trade-off motivates the comparison of \textbf{Strict}, \textbf{Highrecall}, and \textbf{All} strategies in Sec. ~\ref{sec:rq2}.

\begin{table}[t]
\centering
\setlength{\tabcolsep}{6pt}
\caption{Human validation of the LENS static graph (Fleiss' $\kappa{=}0.58$ three-way). Sample composition and definitions of \textbf{Strict} and \textbf{Relaxed} precision are given in \ref{sec:human-validation}; subset rows refer to the Highrecall graph.}
\label{tab:human_val}
\begin{tabular}{lccc}
\toprule
\textbf{Aggregation} & \textbf{\# Triples} & \textbf{Relaxed} & \textbf{Strict} \\
\midrule
Single-run                     & 30 & 83.3\%          & 56.7\% \\
Highrecall (LENS)              & 90 & \textbf{91.1\%} & 54.4\% \\
\midrule
\multicolumn{4}{l}{\textit{Highrecall: subset breakdowns}} \\
\quad $f(e)\geq 0.6$ subset    & 28 & 92.9\%          & 78.6\% \\
\quad non-competitor relations & 52 & 96.2\%          & 63.5\% \\
\midrule
\multicolumn{4}{l}{\textit{Highrecall: by reliability tercile }} \\
\quad Low ($\rho\in[0.16,0.31]$)    & 30 & 90.0\% & 36.7\% \\
\quad Medium ($\rho\in[0.31,0.55]$) & 30 & 90.0\% & 50.0\% \\
\quad High ($\rho\in[0.56,0.96]$)   & 30 & 93.3\% & 76.7\% \\
\bottomrule
\end{tabular}
\end{table}

\begin{table}[t]
\centering
\setlength{\tabcolsep}{5pt}
\caption{Ablation on static graph design (RQ2), S\&P 500, mean over 10 seeds. \textbf{No reliability weight} uses the \emph{same} Highrecall edge set as LENS but replaces $\rho(e)$ with $1$.}

\label{tab:rq2}
\begin{tabular}{lccc}
\toprule
\textbf{Ablation} & \textbf{MRR} $\uparrow$ & \textbf{AR@5} $\uparrow$ & \textbf{SR@5} $\uparrow$ \\
\midrule
\multicolumn{4}{l}{\textit{Graph Source}} \\
\midrule
w/o Static Graph & 0.0361 & 0.3375 & 1.2143 \\
GICS & 0.0354 & 0.3724 & 1.3937 \\
WikiData & 0.0351 & 0.3649 & 1.4237 \\
News Co-occurrence (static) & 0.0370 & 0.3582 & 1.3670 \\
\midrule
\multicolumn{4}{l}{\textit{Construction}} \\
\midrule
Single-run & 0.0371 & 0.3338 & 1.2761 \\
No reliability weight & 0.0369 & 0.4647 & 1.7099 \\
Strict & 0.0370 & 0.4589 & 1.6955 \\
All & 0.0377 & 0.3740 & 1.4139 \\
Highrecall & 0.0371 & 0.5232 & 1.8694 \\
\bottomrule
\end{tabular}
\end{table}

\begin{table}[t]
\centering
\setlength{\tabcolsep}{13pt}
\caption{Static prior graph statistics across aggregation modes (S\&P 500).}
\label{tab:static_graph_stats}
\begin{tabular}{lccc}
\toprule
 & \textbf{Strict} & \textbf{Highrecall} & \textbf{All} \\
\midrule
\#Edges & 3{,}162 & 5{,}424 & 10{,}627 \\
Mean $\rho(e)$ & 0.67 & 0.49 & 0.32 \\
\bottomrule
\end{tabular}
\end{table}

\subsection{Ablation on Static Graph Design (RQ2)}
\label{sec:rq2}

Table~\ref{tab:rq2} examines the impact of static graph source and construction strategies on downstream ranking and portfolio performance. Read together with the human-validation results in Section~\ref{sec:human-validation}, this ablation lets us check whether the monotonic precision patterns observed there carry over to downstream prediction. For ablation studies, we report AR@5 and SR@5 as representative metrics for financial performance.
For GICS and WikiData, we follow the same extraction process as RSR-I \cite{feng2019rsr} for S\&P~500 companies. For news co-occurrence, we leverage the assetId provided in TRNA and build a static graph from the co-occurrence statistics during the training period.

We can observe that LLM-extracted priors (Highrecall) consistently outperform alternative sources. In particular, replacing the static prior with GICS, WikiData, or news co-occurrence graphs leads to clear drops in portfolio metrics (AR@5/SR@5 decrease from 0.5232/1.8694 to 0.3724/1.3937, 0.3649/1.4237, and 0.3582/1.3670, respectively), suggesting that LLM priors provide more informative and complementary cross-company structure than commonly used predefined or heuristic graphs.

We next compare the three aggregation strategies. \textbf{All}, \textbf{Highrecall}, and \textbf{Strict} differ in both edge-set composition and density (Table~\ref{tab:static_graph_stats}: 10{,}627 / 5{,}424 / 3{,}162 edges, mean $\rho=0.32 / 0.49 / 0.67$), so the AR@5/SR@5 ordering 0.3740/1.4139, 0.5232/1.8694, 0.4589/1.6955 reflects a recall-density trade-off rather than an isolated test of continuous weighting. The $f(e)\geq 0.6$ subset used by \textbf{Strict} attains a higher per-edge relaxed precision of 92.9\% in the human validation but discards lower-frequency edges that still encode useful cross-company structure; \textbf{Highrecall} keeps these lower-frequency edges and lets $\rho(e)$ proportionally reduce their influence, which is more expressive than hard thresholding.

To isolate the contribution of continuous $\rho$-weighting from the choice of edge set, the \textbf{No reliability weight} row keeps the Highrecall edge set fixed but only replaces $\rho(e)$ with $1$. Under this matched comparison, removing $\rho$ alone degrades AR@5/SR@5 from 0.5232/1.8694 to 0.4647/1.7099, supporting that propagating $\rho$ as a continuous edge weight is meaningfully better than treating LLM-extracted edges as binary facts.
\textbf{Single-run} uses only one LLM extraction to build the static graph. Compared with multi-run aggregation, it is less stable and yields substantially worse portfolio performance (0.3338/1.2761 vs. 0.5232/1.8694). This is consistent with its lower 83.3\% relaxed precision in Table~\ref{tab:human_val}.

\subsection{Ablation on Dynamic Graph Design (RQ3)}
\label{sec:rq3}

\begin{figure}[t]
    \centering
    \includegraphics[width=\linewidth]{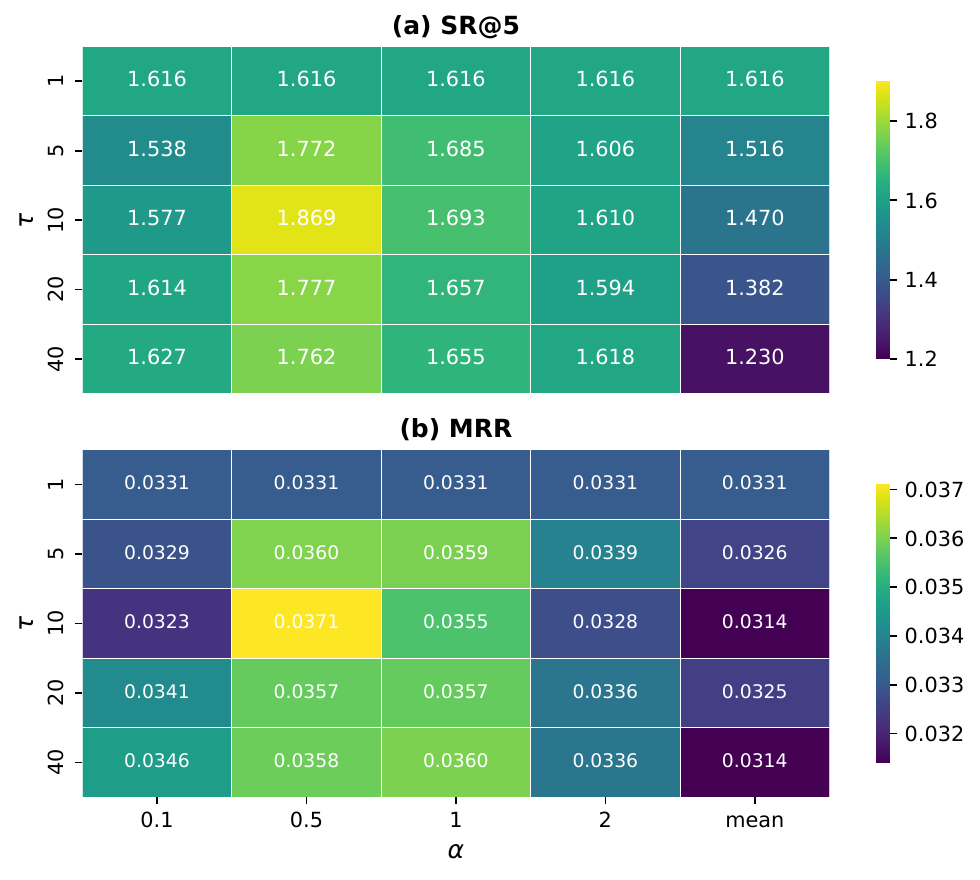}
  \caption{Heatmaps of (a) SR@5 and (b) MRR under different temporal aggregation settings ($\tau \in \{1,5,10,20,40\}$, $\alpha \in \{0.1,0.5,1,2\}$ and mean w/o decay).}
  \Description{Two heatmaps compare SR at 5 and MRR across five temporal window lengths and four decay rates, plus mean aggregation without decay. Both metrics reach their highest values with a 10-day window and a decay rate of 0.5.}
  \label{fig:rq3_heatmap}
\end{figure}

We investigate how the temporal window length $\tau$ and decay rate $\alpha$ affect the news-derived dynamic semantic graph. Figure~\ref{fig:rq3_heatmap} shows that short windows (e.g., $\tau=1$) result in the model being highly sensitive to daily fluctuations, while longer windows (e.g., $\tau=40$) dilute recent signal with outdated information. A moderate window length (approximately $\tau=10$) yields the most stable performance.

The decay rate $\alpha$ determines how strongly the historical information is discounted. Small $\alpha$ keeps outdated information (resembling ``mean'' aggregation), while large $\alpha$ collapses the window to recent news. For example, varing $\tau$ has little effect at $\alpha=2$.
Empirical results indicate that a moderate $\alpha$ (0.5) yields optimal performance, which supports our default setting of $(\tau,\alpha)=(10,0.5)$.

We further evaluate the dynamic graph by substituting it with two alternative sources: historical price (as in GERU \cite{tian2023geru}) and news co-occurrence (as in COGRASP \cite{li2025cograsp}). Table~\ref{tab:ablation_dynamic_graph} shows that the dynamic semantic component is essential: removing it reduces MRR from 0.0371 to 0.0267 and AR@5/SR@5 from 0.5232/1.8694 to 0.2617/1.0854.
While price-based and co-occurrence-based dynamic graphs improve over the no-dynamic variant (e.g., AR@5 / SR@5 of 0.3881/1.4627 and 0.4238/1.5430), they remain below the semantic graphs, indicating that headlines provide meaningful signals beyond static priors.

\begin{table}[t]
\caption{Ablation on dynamic graph design (RQ3). We report the mean performance over 10 seeds.}
\label{tab:ablation_dynamic_graph}
\centering
\setlength{\tabcolsep}{5pt}
\begin{tabular}{lccc}
\toprule
Dynamic graph & \textbf{MRR} $\uparrow$ & \textbf{AR@5}$\uparrow$ & \textbf{SR@5} $\uparrow$ \\
\midrule
Price              & 0.0342 & 0.3881 & 1.4627 \\
News co-occurrence (dynamic) & 0.0342 & 0.4238 & 1.5430 \\
w/o dynamic graph  & 0.0267 & 0.2617 & 1.0854 \\
LENS (LLM static)  & 0.0371 & 0.5232 & 1.8694 \\
\bottomrule
\end{tabular}
\end{table}

\subsection{Ablation on Fusion Strategy (RQ4)}
\label{sec:rq4}

\begin{table}[t]
\centering
\setlength{\tabcolsep}{12pt}
\caption{RQ4: ablations on reliability-aware fusion. Upper panel varies the static/dynamic fusion module; lower panel varies the edge-feature injection in the static GNN (Section~\ref{sec:dual-gnn}). All other components are held fixed. We report the mean performance over 10 seeds.}
\label{tab:rq4_fusion_ablation}
\begin{tabular}{lccc}
\toprule
\textbf{Variant} & \textbf{MRR} $\uparrow$ & \textbf{AR@5} $\uparrow$ & \textbf{SR@5} $\uparrow$  \\
\midrule
\multicolumn{4}{l}{\textit{Fusion strategy}} \\
\midrule
Equal (0.5/0.5) & 0.0354 & 0.4361 & 1.6278 \\
Concat (MLP) & 0.0312 & 0.3608 & 1.3756 \\
Learned & 0.0371 & 0.4246 & 1.5899 \\
Gated (LENS) & 0.0371 & 0.5232 & 1.8694 \\
\midrule
\multicolumn{4}{l}{\textit{Edge-feature injection (static GNN)}} \\
\midrule
Key only     & 0.0367 & 0.4999 & 1.8136 \\
Value only   & 0.0356 & 0.5060 & 1.8302 \\
Dual (LENS)  & \textbf{0.0371} & \textbf{0.5232} & \textbf{1.8694} \\
\bottomrule
\end{tabular}
\end{table}
\begin{figure}[t]
\centering
\includegraphics[width=\linewidth]{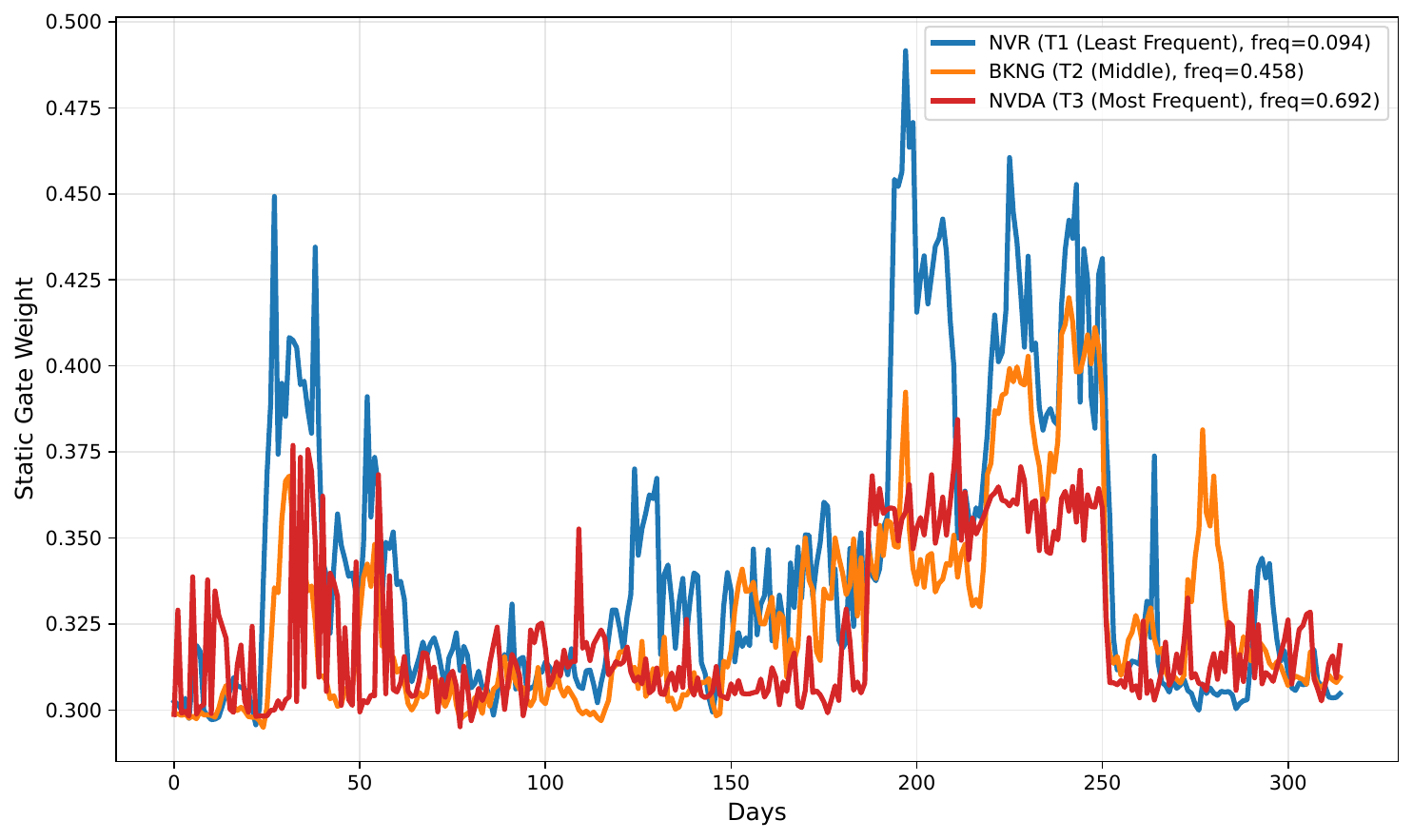}
\caption{Static gate weights for representative stocks from different news coverage groups: NVR (low coverage, freq=0.094), BKNG (medium coverage, freq=0.458), and NVDA (high coverage, freq=0.692).}
\Description{Line chart of static gate weights over approximately 320 trading days for NVR, BKNG, and NVDA, representing low, medium, and high news coverage. NVR generally receives higher static weights, and all three stocks show increased weights around day 190.}
\label{fig:gate_weights}
\end{figure}
\begin{table}[t]
\centering
\setlength{\tabcolsep}{5pt}
\caption{MAE across different news coverage groups. We report the mean performance over 10 seeds.}
\label{tab:mae_tier}
\begin{tabular}{lcccc}
\toprule
\textbf{Method} & \textbf{High} & \textbf{Medium} & \textbf{Low} & \textbf{$\Delta$(H→L)} \\
\midrule
Dynamic Only & 0.012690 & 0.012793 & 0.013374 & 5.39\% \\
LENS (Equal) & 0.012738 & 0.012768 & 0.013279 & 4.24\% \\
LENS (Gated) & 0.012752 & 0.012725 & 0.013158 & 3.18\% \\
\bottomrule
\end{tabular}
\end{table}

To answer RQ4, we ablate two reliability-aware design choices: gated fusion (Section~\ref{sec:fusion}) rebalances static and dynamic branches per stock and day, while dual edge-feature injection (Section~\ref{sec:dual-gnn}) propagates $\rho$ through attention logits and message values in the static GNN. Either stage could lose the reliability signal.

We compare our gated fusion against three baseline strategies:
(i) \textbf{Concat}: static and dynamic representations are concatenated and projected through a linear layer;
(ii) \textbf{Equal}: the two representations are combined with fixed equal weights;
(iii) \textbf{Learned}: a single static weight is learned and shared across all stocks and time steps.
In contrast, our gated fusion computes context-dependent weights per stock per day.
Table~\ref{tab:rq4_fusion_ablation} (upper panel) shows Gated fusion achieves the best AR@5 and SR@5.
Concat performs the worst, suggesting that implicit fusion via linear projection is less effective than even the coarse explicit combination of Equal weighting. Notably, Learned and Gated achieve identical MRR, yet Gated surpasses Learned by 23.2\% in AR@5 (0.5232 vs. 0.4246) and 17.6\% in SR@5 (1.8694 vs. 1.5899), showing the value of context-dependent weighting: a scalar cannot adapt to per-stock, per-day variation in news coverage.

The lower panel of Table~\ref{tab:rq4_fusion_ablation} ablates the dual edge-feature injection in the static GNN. Restricting injection to the key alone (which neighbors receive attention) or to the value alone (what is aggregated) underperforms the dual variant: AR@5 falls from 0.5232 to 0.4999 (key-only) and 0.5060 (value-only), with corresponding drops in SR@5. The two pathways carry complementary information and collapsing either truncates the reliability signal before it reaches the node representation.

We also examine the learned gate behaviour under different news coverage.
Figure~\ref{fig:gate_weights} visualises the static gate weight for representative stocks with low/medium/high news coverage.
For the low-coverage stock (NVR), the model assigns higher static weights, effectively falling back to the stable LLM prior when dynamic evidence is weak; for a high-coverage stock (NVDA), the gate becomes more responsive and assigns lower static weights when the news signal is rich. The medium-coverage stock (BKNG) falls between the two extremes.
Additionally, all three stocks show elevated weights around Day 190 (a period of market turbulence). These results illustrate that the gated fusion mechanism adapts at both the stock level and the market level.

Furthermore, Table~\ref{tab:mae_tier} reports MAE across news coverage groups. All methods show higher prediction error for low-coverage stocks.
But LENS is more robust: Dynamic Only shows a 5.39\% MAE increase from High to Low tier, while LENS (Gated) reduces this degradation to 3.18\%. The improvement can be attributed to both components. Adding static priors (LENS Equal vs. Dynamic Only) reduces the gap by 1.15\%, indicating that LLM-extracted relations provide stable signals when news is sparse. Switching to adaptive fusion (LENS Gated vs. Equal) further reduces the gap by 1.06\%, indicating that context-dependent weighting effectively leverages static priors for under-covered stocks, which is consistent with the gate weight patterns observed in Figure~\ref{fig:gate_weights}.

\section{Conclusions}
Graph-based stock prediction methods use relational signals from curated knowledge bases, industry taxonomies, news co-mentions or price correlations whose reliability is uncertain.
LENS carries the reliability of heterogeneous evidence through the pipeline rather than simplifying it at one stage. 
The framework keeps LLM-extracted edges weighted by multi-run aggregation, builds news-driven semantic graphs with temporal decay to resist single-day noise, and learns a per-stock, per-day gate that rebalances the two as news coverage varies. 
On S\&P 500 and NASDAQ 100, this design improves ranking and portfolio metrics over strong sequence- and graph-based baselines. A small-sample human-validation study indicates that most extracted edges describe real corporate relationships.
The ablation studies show how each part of the reliability-aware framework contributes.

Three scope boundaries deserve mention. First, the static graph is extracted from a single LLM whose knowledge cutoff precedes our evaluation horizon, which controls information leakage but leaves cross-extractor robustness untested. Second, both evaluation universes are US large-cap, so the framework's behaviour in markets where company-level relations are sparser, or in regimes that are not primarily bullish, remains open. Third, LENS is evaluated through historical backtests rather than live deployment, and the comparison covers published academic baselines rather than proprietary industry ensembles. These limits frame the natural extensions: cross-LLM extractor experiments, multi-horizon prediction in non-US markets, deployment-aware evaluation, and finer-grained event-driven dynamic graphs.

\newpage

\section*{GenAI Usage Disclosure}

Generative AI tools were used in this work for two purposes. First, as described in the paper, large language models were used as part of the proposed relation extraction pipeline to identify company relationship candidates from textual/semantic inputs. The extracted relations were subsequently processed, filtered, and evaluated according to the methodology described in the paper. Second, GenAI tools were used to assist with language polishing, grammar correction, and improving the clarity of the manuscript. 

GenAI tools were not used to generate or modify the reported experimental results, to implement the prediction model, to write or execute the training/evaluation code, or to perform the non-LLM data preprocessing and numerical analysis. All technical claims, experimental results, interpretations, and final manuscript content were reviewed and verified by the authors, who take full responsibility for the submitted work.

\bibliographystyle{ACM-Reference-Format}
\bibliography{sample-base}

\end{document}